\documentclass{article}

\usepackage[preprint]{neurips_2025}

\usepackage[utf8]{inputenc} 
\usepackage[T1]{fontenc}    
\usepackage{hyperref}       
\usepackage{url}            
\usepackage{booktabs}       
\usepackage{amsfonts}       
\usepackage{nicefrac}       
\usepackage{microtype}      
\usepackage{xcolor}         
\usepackage{enumitem}
\usepackage{amsmath} 
\usepackage{graphicx}
\usepackage{wrapfig}

\title{Learning from Less: Guiding Deep Reinforcement Learning with Differentiable Symbolic Planning}

\author{%
  Zihan Ye$^{1,2}$ \quad
Oleg Arenz$^{1}$\quad
Kristian Kersting$^{1,2,3,4}$ \\
$^{1}$Computer Science Department, TU Darmstadt, Germany \\
$^{2}$Hessian Center for Artificial Intelligence (hessian.AI), Darmstadt, Germany \\
$^{3}$Centre for Cognitive Science, TU Darmstadt, Germany \\
$^{4}$German Research Center for Artificial Intelligence (DFKI), Darmstadt, Germany \\
\texttt{\{firstname.lastname\}@tu-darmstadt.de}
}

\begin{document}

\maketitle
\begin{abstract}
When tackling complex problems, humans naturally break them down into smaller, manageable subtasks and adjust their initial plans based on observations. For instance, if you want to make coffee at a friend’s place, you might initially plan to grab coffee beans, go to the coffee machine, and pour them into the machine. Upon noticing that the machine is full, you would skip the initial steps and proceed directly to brewing. In stark contrast, state-of-the-art reinforcement learners, such as Proximal Policy Optimization (PPO), lack such prior knowledge and therefore require significantly more training steps to exhibit comparable adaptive behavior. Thus, a central research question arises: \textit{How can we enable reinforcement learning (RL) agents to have similar ``human priors'', allowing the agent to learn with fewer training interactions?} To address this challenge, we propose \textbf{d}ifferentiable s\textbf{y}mbolic p\textbf{lan}ner (Dylan), a novel framework that integrates symbolic planning into Reinforcement Learning. Dylan serves as a reward model that dynamically shapes rewards by leveraging human priors, guiding agents through intermediate subtasks, thus enabling more efficient exploration. Beyond reward shaping, Dylan can work as a high-level planner that composes primitive policies to generate new behaviors while avoiding common symbolic planner pitfalls such as infinite execution loops. Our experimental evaluations demonstrate that Dylan significantly improves RL agents' performance and facilitates generalization to unseen tasks.
\end{abstract}

\section{Introduction}
Reinforcement learning (RL) has demonstrated remarkable success across a wide range of domains, including game playing \citep{mnih2015human,silver2021learning}, robotic manipulation \citep{pmlr-v100-neunert20a,andrychowicz2020learning}, and more recently, post-training in Large Language Models (LLMs) \citep{ouyang2022training,liu2024deepseek,liu20242deepseek,guo2025deepseek}. Despite these advances, the challenge of sparse rewards remains a significant barrier to the broader applicability and efficiency of RL methods~\citep{ng1999policy,andrychowicz2017hindsight,ecoffet2021first}. In sparse environments, the reward signals are infrequent or delayed, making effective exploration difficult, and the learning process can become prohibitively expensive in terms of computation \citep{pathak2017curiosity}. Furthermore, sparse rewards often lack the granularity needed to explicitly guide agents toward desirable behaviors, frequently resulting in suboptimal or unintended outcomes. For example, while LLMs such as DeepSeek-R1 demonstrate a certain level of reasoning ability in solving complex problems, they may still exhibit issues such as language mixing or inconsistent linguistic patterns during multi-step reasoning—largely due to the limited structure provided by sparse rewards during post-training \citep{guo2025deepseek}. Similarly, in robotics, sparse reward signals can lead agents to exploit unintended shortcuts or converge on behaviors that diverge from human expectations \citep{ng1999policy,christiano2017deep}.

Prior works~\citep{jaderberg2017reinforcement,ng1999policy} have explored reward shaping as a means of providing agents with additional learning signals to address this challenge. These signals are typically derived from potential-based functions \citep{Ng1999PolicyIU}, expert-crafted heuristics \citep{Kober2013ReinforcementLI, Devlin2011AnES}, or learned reward models based on human preferences \citep{christiano2023deepreinforcementlearninghuman} and auxiliary tasks \citep{jaderberg2017reinforcement}. 
While these approaches can reduce the amount of data required for learning to some extent, a fundamental question remains: \textit{Can we design a reward model that not only enables agents to learn with fewer training interactions, but also aligns with human intent and remains inherently interpretable?}

To address this question, we propose \textsc{Dylan}, a differentiable symbolic planner that serves as a reward model to guide reinforcement learning agents. Unlike prior reward models, \textsc{Dylan} decomposes the goal task into modular subgoals and assigns rewards through logical reasoning over these subgoals (e.g., \textit{the coffee beans are at hand}, \textit{the agent is at the coffee machine}). These subgoal-based rewards are semantically aligned with human understanding of task structure, allowing agents to learn more efficiently while preserving interpretability.
Beyond its role as a reward model, \textsc{Dylan} can also work as a symbolic planner that composes different behaviors by stitching together reusable policy primitives. This capability alleviates a key limitation of conventional RL agents: they are overfit to a single task and fail to generalize. For instance, an agent trained to \textit{navigate to a red door} may perform poorly when asked to \textit{pick up a blue key and open a blue door}. Without modular reasoning and task composition, current RL systems must be retrained for every new task, incurring significant cost. \textsc{Dylan}'s compositional structure supports generalization across related tasks and facilitates knowledge transfer through its symbolic task grounding. Moreover, \textsc{Dylan}'s differentiable nature overcomes a common limitation of traditional symbolic planners, which are typically non-adaptive and prone to failure in environments requiring flexible search strategies.

Overall, we make the two following major, novel contributions:
\begin{itemize}[leftmargin=20pt, itemsep=0pt,parsep=0pt,topsep=-3pt,partopsep=0pt]
    \item We introduce \textsc{Dylan}, a differentiable symbolic planner that can be integrated as a reward model into existing reinforcement learning frameworks, offering interpretable intermediate feedback that guides agents with fewer interactions while remaining aligned with human intent.
    
    
    \item Beyond reward model, \textsc{Dylan} can also serve as a differentiable planner. Acting as a high-level policy in a hierarchical RL setting, it composes policy primitives in a modular and flexible way, allowing the agent to generate new behaviors.
\end{itemize}


To this end, we proceed as follows. We start off by discussing background and related work in Sec.~\ref{background}, followed by a detailed introduction of \textsc{Dylan} in Sec.~\ref{dylan}. Before concluding, we touch upon the experimental section in Sec.~\ref{experiments}.

\begin{figure}[t]
    \centering
    \includegraphics[trim=75 255 141 21, clip, width=\columnwidth]{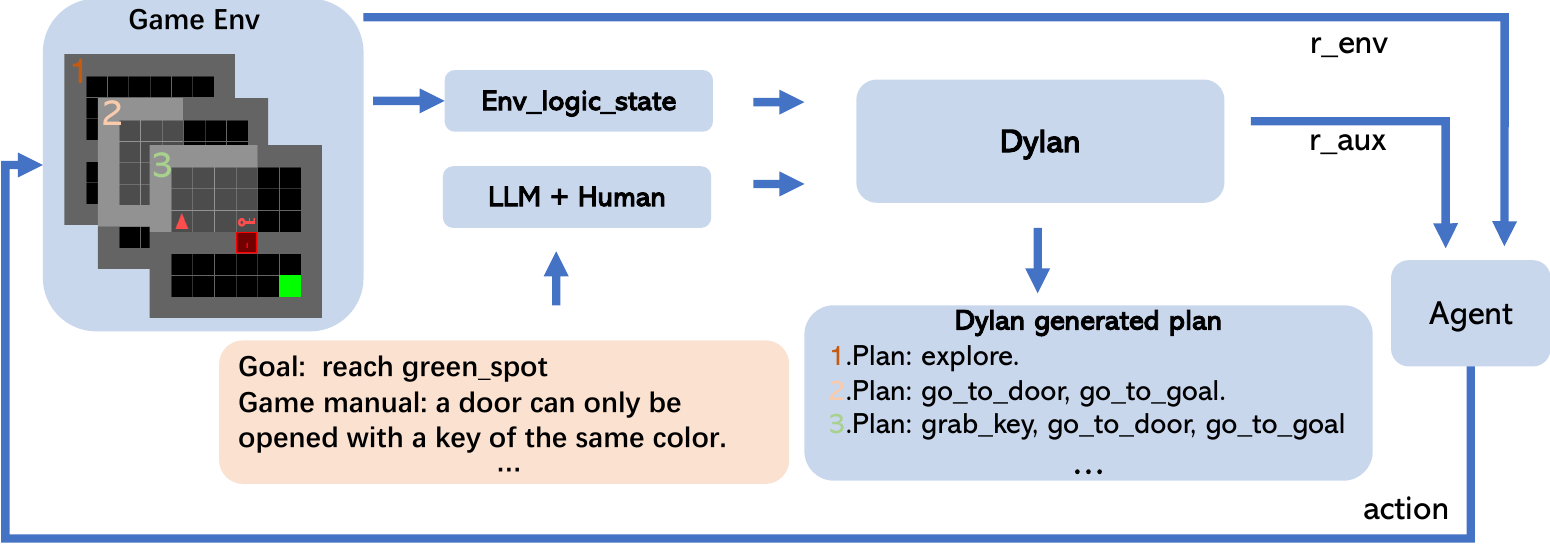}
    \caption{\textbf{An overview of Dylan working as a reward model.} Given the goal and game rules, Dylan first generates candidate plans to achieve the goal and then shapes rewards based on the generated plans, balancing among all candidate plans or selecting the best one to follow through.} 
    \label{fig:rewardmodel}
\end{figure}

\section{Background and Related Work}
\label{background}
\textsc{Dylan} builds upon several research areas, including first-order logic, differentiable reasoning, reinforcement learning and reward shaping. 

\textbf{First-order logic and Differentiable Forward-Chaining Reasoning.} 
We refer readers to Appendix~\ref{FOL} for a review of first-order logic fundamentals. Build upon this foundation, differentiable forward-chaining inference~\citep{Evans18,Shindo21aaai,ye2022differentiable} enables logical entailment to be computed in a differentiable manner using tensor-based operations. This technique bridges symbolic reasoning with gradient based learning, allowing logic driven models to be trained end-to-end. Building on this foundation, a variety of extensions have emerged, including reinforcement learning agents that incorporate logical structures into policy learning~\citep{jiang2019neural,delfosse2023interpretable}, differentiable rule learners capable of extracting interpretable knowledge from complex visual environments~\citep{shindo2023alpha} and  differentiable meta-reasoning frameworks that enable a reasoner to perform self-inspection~\citep{ye2022differentiable}.

Dylan builds upon these developments by introducing differentiable meta-level reasoning within the forward-chaining framework~\citep{shindo2023alpha}, thereby enabling symbolic planning capabilities in a fully differentiable architecture. Unlike classical symbolic planners~\citep{fikes1971strips,fikes1993strips}, Dylan learns to adaptively compose and select rules based on task demands, thus alleviating the non-adaptivity limitation of previous classical symbolic planners.


\textbf{Deep Reinforcement Learning.}
Reinforcement Learning (RL) has been extensively studied as a framework for sequential decision-making, where an agent learns an optimal policy through trial and error. Traditional RL approaches, such as Q-learning \citep{watkins1992q} and policy gradient methods \citep{sutton2000comparing}, have been foundational to modern RL advancements. With the rise of deep learning, Deep Reinforcement Learning (DRL) has demonstrated success in high-dimensional environments, particularly in Advantage Actor-Critic (A2C) \citep{mnih2016asynchronous} and Proximal Policy Optimization (PPO) \citep{schulman2017proximal}. However, standard RL methods often struggle with sparse or delayed rewards, leading to inefficient exploration. To address this, intrinsic motivation \citep{pathak2017curiosity} and curriculum learning \citep{bengio2009curriculum} have been proposed to encourage exploration and guide policy learning in complex environments. Despite significant progress, RL algorithms still suffer from sample inefficiency, reward sparsity, and generalization issues. Recent research focuses on improving stability and efficiency through techniques such as hierarchical RL~\citep{nachum2018data}, meta-learning~\citep{finn2017model} and model-based RL \citep{kaisermodel}. 

Dylan builds upon these advancements by serving as a reward model, aiming to enhance learning efficiency. Unlike traditional model-based reinforcement learning \cite{moerland2023model,wu2022plan}, the objective of Dylan is not to construct a world model via the logic planner. Instead, Dylan leverages human prior knowledge encoded in the logic planner to shape the agent's behavior through informed reward signals, thus guiding and accelerating the learning process, rather than modeling environment dynamics. 

\textbf{Reward Shaping.}
Reward shaping has been extensively explored to improve sample efficiency and learning stability in RL, particularly under sparse or delayed feedback. Classical potential-based reward shaping methods \citep{ng1999policy} preserve policy invariance by incorporating scalar potential functions into the reward signal. While this method is theoretically well-grounded, their expressiveness is limited. This limitation arises from their reliance on hand-crafted heuristics and their inability to capture complex task structures. Potential-based shaping has been applied to multi-agent systems \citep{Devlin2011AnES}, showcasing its effectiveness in cooperative settings through the use of spatial and role-specific heuristics. However, the approach still lacks compositionality and offers limited interpretability. To address the limitations of heuristic-driven shaping, recent approaches have explored unsupervised learning signals as an alternative means of guiding agent behavior. Unsupervised auxiliary tasks~\citep{jaderberg2017reinforcement}, generate pseudo-rewards to facilitate representation learning and encourage exploration. However, the resulting shaping signals are typically task-agnostic, static, and misaligned with high-level semantic objectives. To overcome the limitations of generic auxiliary signals, subsequent work has explored leveraging human preferences to ground reward learning in semantically meaningful objectives. Preference-based reward learning \citep{christiano2023deepreinforcementlearninghuman} aligns the agent's behavior with human preference by inferring reward functions from trajectory comparisons. While effective, this method suffers from poor sample efficiency and produces reward models that are often opaque and difficult to interpret or generalize.


Unlike existing reward shaping approaches that employ static heuristics or task-agnostic signals, Dylan introduces a differentiable symbolic planner as the reward model. This enables the dynamic assignment of shaped rewards that reflects task structure. Being 
interpretable and modular, 
Dylan facilitates learning through subgoal decomposition and compositional planning.

\section{Dylan: Guiding RL using Differentiable Symbolic Planning}
\label{dylan}
\begin{figure}[t]
    \centering
    \includegraphics[trim=96 70 32 71, clip, width=\columnwidth]{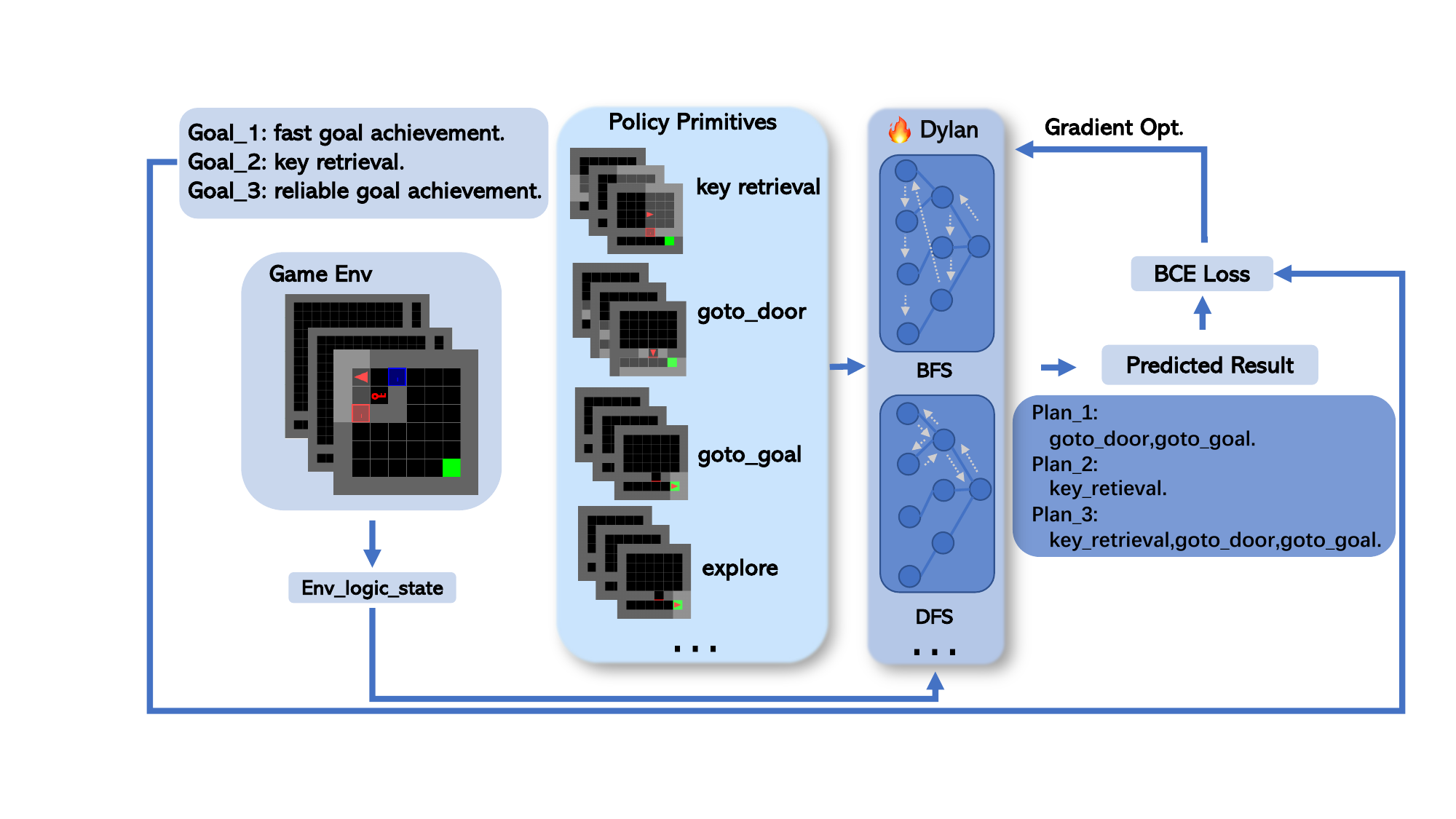}
    \caption{\textbf{Dylan, as a differentiable symbolic planner}, is capable of adapting to tasks that require different search strategies and stitching together different primitive policies to generate novel behaviours. (Best viewed in color)}
    \label{fig:plannermodel}
\end{figure}
A key feature of Dylan is its ability to incorporate human prior knowledge into (deep) RL in a structured and actionable end-to-end fashion. To this end, Dylan uses (differentiable) symbolic logic to represent prior knowledge and guide the agent’s through reward shaping. Let us illustrate this using a navigation task from the MiniGrid-DoorKey environment~\citep{chevalier2023minigrid}, where the agent must pick up a key, unlock a door, and reach the goal.  By consulting the environment’s manual, we extract high-level rules like: “a door can only be opened with a key of the same color” and “the agent can reach the goal only by passing through an open door.” Instead of representing such rules as raw text or hardcoded logic, Dylan uses them as structured symbolic transitions: each step describes an action (e.g., go to the key) that transforms one state (e.g., no key) into another (e.g., has key), provided certain preconditions are met. These transitions are represented in a format inspired by STRIPS~\citep{fikes1971strips}, commonly used in classical planning. This symbolic abstraction allows Dylan to build a high-level plan that sequences primitive actions to achieve a given goal. To obtain task-specific rules as structured symbolic transitions, we prompt GPT-4o~\citep{openai_chatgpt4o_2024} to generate and transform candidate game rules, and then rely on human supervision to verify and refine them. We provide the detailed environment logic, game rules, the logic planner and the prompt format in Appendix~\ref{prompt}.
Akin to \cite{Shindo21aaai, shindo2023alpha}, by defining the initial and $t$-th step valuation of ground atoms as $\mathbf{v}^{(0)}$ and $\mathbf{v}^{(t)}$, we make the symbolic planner differentiable in three steps: \textbf{(Step 1)} We encode each planning rule \( C_i \in \mathcal{C} \) as a tensor \( \mathbf{I}_i \in \mathbb{N}^{G \times S \times L} \), where \( S \) is the maximum number of possible substitutions for variables, \( L \) is the maximum number of body atoms and \( G \) is the number of grounded atoms. Specifically, the tensor \( \mathbf{I}_i \) stores at position $[j, k, l]$ the index (0 to $G-1$) of the grounded atom that serves as the $l$-th body atom when rule $C_i$ derives grounded head $j$ using substitution $k$. \textbf{(Step 2)} To be able to learn which rules are most relevant during forward reasoning, a weight matrix $\mathbf{W}$ consisting of $M$ learnable weight vectors, $[\mathbf{w}_1, \dots, \mathbf{w}_M]$, is introduced. Each vector $\mathbf{w}_m \in \mathbb{R}^C$ contains raw weights for the $C$ planning rules. To convert these raw weights into normalized probabilities for soft rule selection, a \textit{softmax} function is applied independently to each vector $\mathbf{w}_m$, yielding $\mathbf{w}^*_m$. \textbf{(Step 3)} At each step \( t \), we compute the valuation of body atoms using the \textit{gather} operation over the valuation vector \( \mathbf{v}^{(t)} \), looping over the body atoms for each grounded rule. These valuations are combined using a soft logical AND (\emph{gather} function) followed by a soft logical OR across substitutions:
\begin{equation}
    b_{i,j,k}^{(t)} = \prod\nolimits_{1 \leq l \leq L} \mathbf{gather}(\mathbf{v}^{(t)}, \mathbf{I}_i)[j,k,l], \quad
    c_{i,j}^{(t)} = \mathit{softor}^\gamma (b_{i,j,1}^{(t)}, \dots, b_{i,j,S}^{(t)}).
\end{equation}

Here, \( i \) indexes the rule, \( j \) the grounded head atom, and \( k \) the substitution applied to existentially quantified variables. The resulting body evaluations \( c_{i,j}^{(t)} \) are weighted by their assigned rule weights \( w^*_{m,i} \), and then aggregated across rules and rule sets:
\begin{equation}
    h_{j,m}^{(t)} = \sum\nolimits_{1 \leq i \leq C} w^*_{m,i} \cdot c_{i,j}^{(t)}, \quad
    r_{j}^{(t)} = \mathit{softor}^\gamma ( h_{j,1}^{(t)}, \dots, h_{j,M}^{(t)} ), \quad
    v^{(t+1)}_j = \mathit{softor}^\gamma (r^{(t)}_j, v^{(t)}_j).
\end{equation}

We provide full details of this differentiable reasoning procedure in Appendix~\ref{diff}.

\subsection{Dylan as Reward Model}
\label{GreedyRewardmodel}
With Dylan at hand, we now incorporate it into the reinforcement learning and use it as a static \textbf{reward model}. 
As the agent interacts with the environment, a logical state is provided by the environment and used as an input to Dylan. Based on this received state, goal and rules, Dylan performs reasoning to identify the most promising plan to reach the goal. Initially, an exploratory plan is executed to gather information about the environment. After this exploratory phase, Dylan selects the optimal \textit{plan}, the one with the highest estimated probability of achieving the desired goal. 
Once the plan is determined, the reward distribution follows the plan during subsequent decision-making.

\begin{figure}[t]
    \centering
    \includegraphics[width=0.495\textwidth]{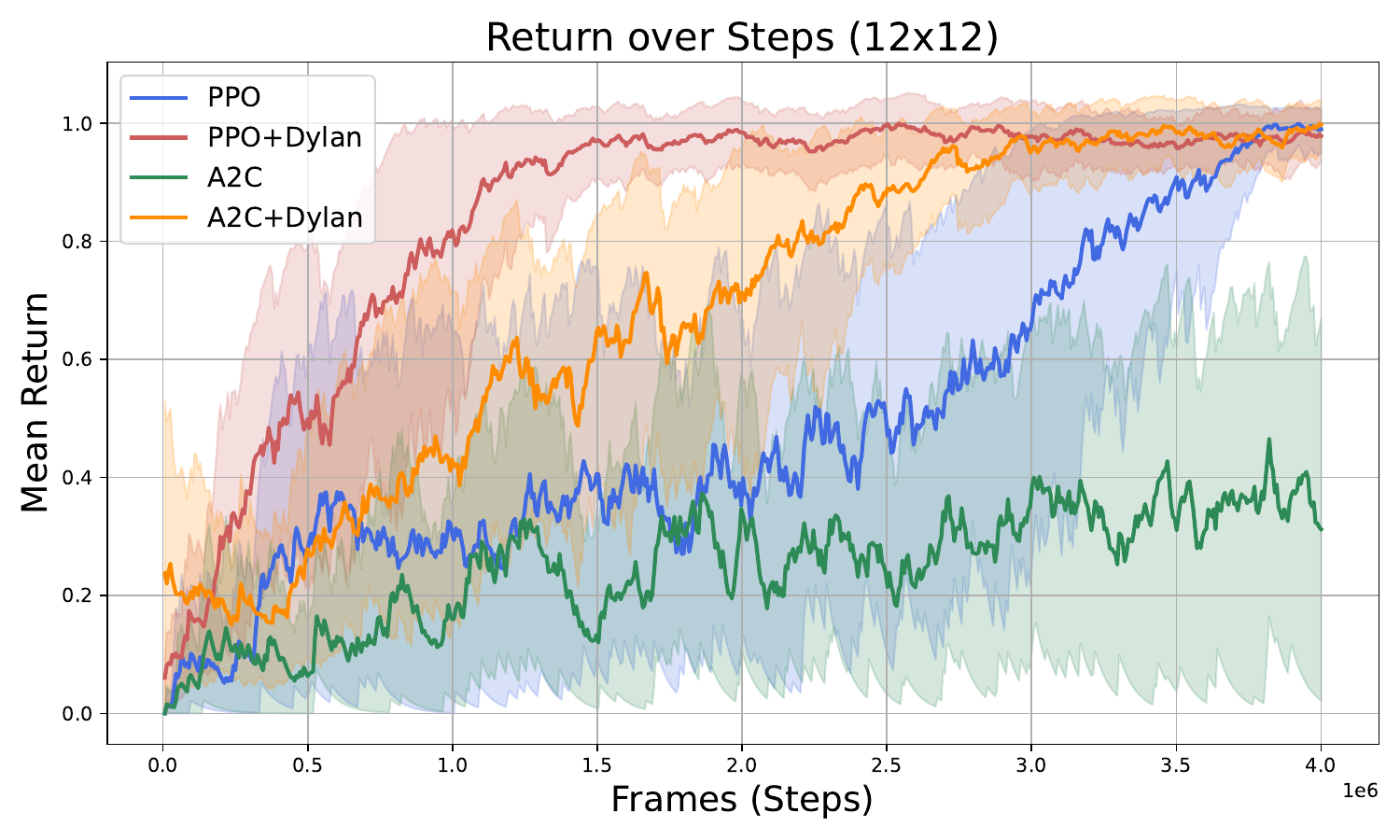}
    \includegraphics[width=0.495\textwidth]{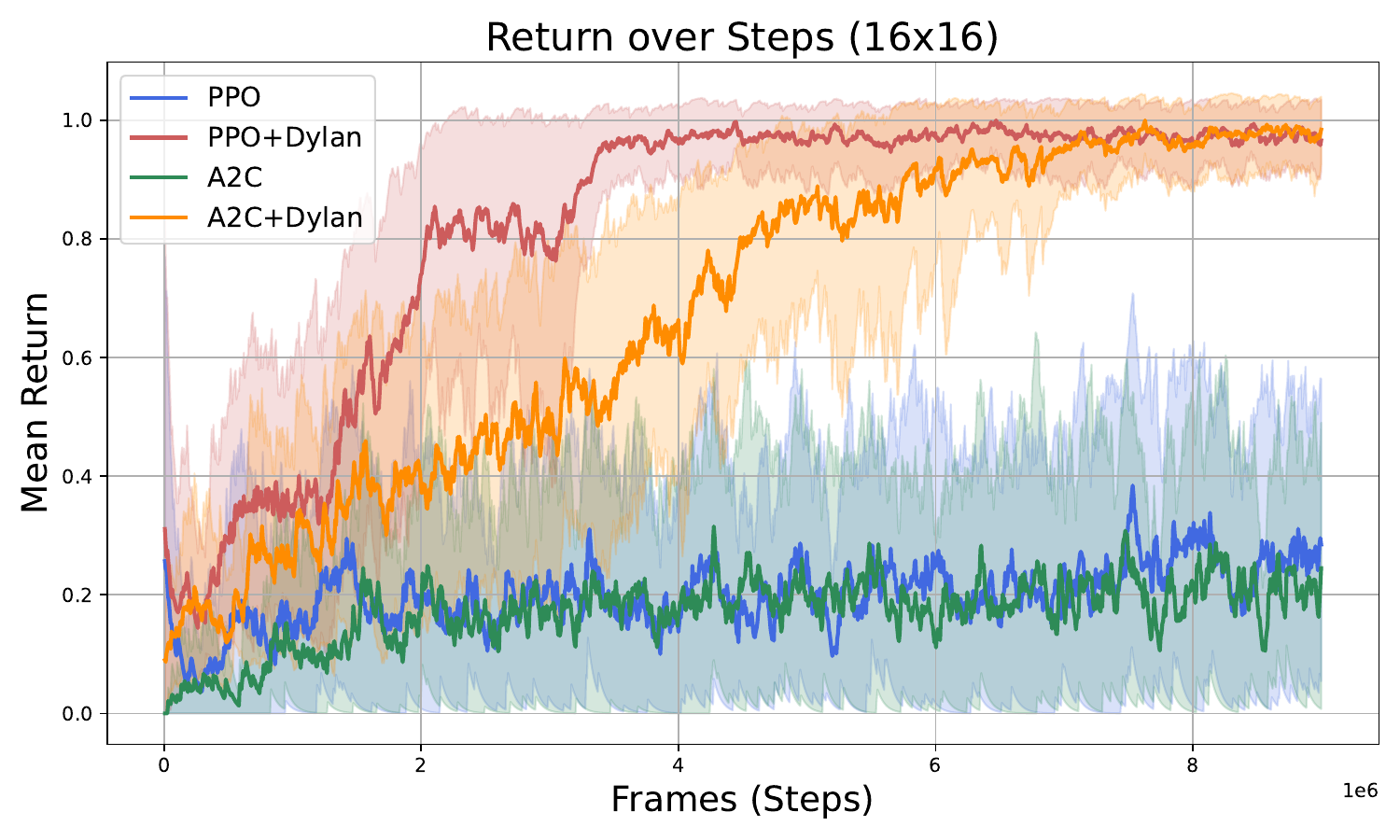}
    \caption{\textbf{Dylan boost
RL agents’ performances as a static reward model.} Return comparisons of methods with and without Dylan in $\mathtt{12\times12}$ and $\mathtt{16\times16}$ MiniGrid environments during training. The return curves are averaged over three runs, with the solid lines representing the mean values and the shaded areas indicating the minimum and maximum values. All curves are smoothed using exponential moving average (EMA) for improved readability. (Best viewed in color)}
    \label{fig:rewardmodel}
    \vspace{-0.2in}
\end{figure}
Let the selected action sequence be denoted as \(\mathtt{[a_1, a_2, a_3, \dots, a_n]}\), where each action \(\mathtt{a_i}\) represents a high-level action. Unlike the low-level actions commonly used in reinforcement learning, these high-level actions encode semantically meaningful behaviors, such as \(\mathtt{get\_key}\). For each action, the planner also receives the corresponding expected state transition, represented as \(\mathtt{move(a_i, s_{\text{pre}}^{(i)}, s_{\text{post}}^{(i)})}\), where \(\mathtt{s_{\text{pre}}^{(i)}}\) and \(\mathtt{s_{\text{post}}^{(i)}}\) denote the pre- and post-action symbolic states, respectively as introduced previously. The \textbf{reward function} is designed to provide feedback only when the agent follows the planned action sequence in the correct order. Specifically, the agent must achieve each planned post-condition state \(\mathtt{s_{post}^{(i)}}\), corresponding to action \(\mathtt{a_i}\), before proceeding to the next action in the sequence. No reward is given if the agent deviates from the prescribed order or reaches states out of sequence. The auxilary reward function \(r_{\text{reasoner}}(s, a, i)\) at 
$\mathtt{i}$-th transition is defined as:
\begin{equation}
r_{\text{reasoner}}(s, a, i) =
\begin{cases} 
    \mathtt{max}\big((\lambda - {\mathtt{num\_steps}}\slash{\mathtt{total\_steps}}), \mathtt{0}\big), & \text{if } s = s_{post}^{(i)}, \\
    \mathtt{0}, & \text{otherwise}.
\end{cases}
\end{equation}
Here $\mathtt{\lambda>0}$, is a hyperparameter which denotes the reward. Dylan ensures the rewards are assigned \textbf{sequentially}, strictly following the planned order of actions \(\mathtt{[a_1, a_2, \dots, a_n]}\).
The agent must complete each transition \(\mathtt{move(a_i, s_{pre}^{(i)}, s_{post}^{(i)})}\) before progressing to the next step $\mathtt{i}$+$\mathtt{1}$. No rewards are granted if the agent skips steps, performs actions out of order, or transitions to incorrect states. Additionally, the reward is penalized by the number of steps taken $\mathtt{num\_steps}$ relative to the total allowed steps $\mathtt{total\_steps}$. This incentivizes the agent to follow the planned sequence as efficiently as possible and discourages unnecessary actions.
The shaped reward is defined as the sum of the environment reward and the reasoner reward:
\begin{equation}
r'(s, a, i) = r(s, a)_{\text{env}} + r_{\text{reasoner}}(s, a, i)
\end{equation}
The objective becomes maximizing the expected cumulative discounted shaped reward:
\begin{equation}
J'(\pi) = \mathbb{E}_{\tau \sim \pi} \left[ \sum\nolimits_{t=0}^{\infty} \gamma^t \left( r(s, a)_{\text{env}} + r_{\text{reasoner}}(s, a, i) \right) \right]
\end{equation}
By integrating the planner as a reward model, we provide structured and goal-aligned shaped rewards. This can lead to faster convergence and fewer training interactions, as shown empirically in the experimental section.
\subsection{Dylan as Adaptive Reward Model}
In this subsection, we use Dylan as an adaptive reward model for reinforcement learning agents, leveraging probabilities of all possible plans from Dylan. 

Building on the reward model described in Sec.~\ref{GreedyRewardmodel}, we introduce an additional dense reward 
$r_{\text{adaptive}}$ for each action. This reward encourages the agent to take steps that more effectively lead toward subgoals, thereby further accelerating learning process. The dense reward $r_{\text{adaptive}}$ is computed using the \emph{log-sum-exp}~\citep{cuturi2017soft} over the probabilities of all candidate plans, ensuring numerical stability while leveraging the full distribution of plan probabilities. To obtain these probabilities, Dylan performs reasoning at each step of the agent’s trajectory, producing updated likelihoods for each plan. The probability of each plan is obtained by
$
    p_{\text{targets}} = \mathbf{v}^{(T)}\left[I_{\mathcal{G}}(\operatorname{targets})\right],$ where   $I_{\mathcal{G}}(x)$ returns the index of the target atom within the set $ \mathcal{G} $.  
 \( \mathbf{v}^{(T)} = f_\text{infer}(\mathbf{v}^{(0)})
\) denotes the valuation tensor obtained after \( T \)-step forward reasoning. \( \mathbf{v}[i] \) represents the \( i \)-th entry of the valuation tensor. Details on how \( \mathbf{v}^{(0)} \) is initialized are provided in Appendix~\ref{v0}.

We aggregate the probabilities of all plans using the \emph{log-sum-exp}~\citep{cuturi2017soft} operation for numerical stability, resulting in the dense reward:
\[
r_{\text{adaptive}} = \log\left(\sum\nolimits_{p_j \in \mathcal{P}_{a_t}^{(i)}} \exp\left(\log (p_j)\right)\right).
\]
To ensure that the agent’s learning is not 
dominated solely by the dense reward, we scale it by a factor $\omega$. As positive dense rewards could discourage the agent from finishing an episode by entering zero-reward absorbing states, we subtract a positive constant $\lambda$ ensuring that the dense auxiliary reward is always negative. The sparse rewards, discussed in Sec.~\ref{GreedyRewardmodel} does not suffer from such survival bias, as it only given when the agent makes progress towards the goal. The resulting reward function, which integrates the adaptive reward with both the environment reward and the reasoner reward described in Sec.~\ref{GreedyRewardmodel}, is defined as:
\[
r'(s, a, i) = 
\begin{cases}
\omega * r_{\text{adaptive}}  -\lambda+ r(s, a)_{\text{env}}+ r_{\text{reasoner}}(s, a, i), & \text{successful transition}, \\[6pt]
\omega * r_{\text{adaptive}}-\lambda+ r(s, a)_{\text{env}}, & \text{otherwise}.
\end{cases}
\]

Instead of sticking to a single fixed plan, our adaptive reward model takes all possible plans into consideration, guiding the agent to choose actions aligned with high-probability, goal-achieving strategies while discouraging stagnation or ineffective behavior. 

\subsection{Dylan as Differentiable Planner}
In addition to serving as a reward model within the reinforcement learning training, Dylan can also operate as a standalone planner, enabling the integration of multiple policies outside the training process. As a planner, Dylan's primary objective is to stitch together diverse primitive policies to generate new behaviors, thus enabling the agent to finish new tasks without retraining. However, combining various policies within limited reasoning steps presents a significant challenge: the risk of becoming trapped in loops or suffering from inefficient exploration strategies. Consider a planning task where the objective is to reach state C from state A, with available transitions including A→B, B→A, and B→C. When using depth-first search (DFS), the planner may enter an infinite loop, repeatedly cycling through states without reaching the goal.
For instance, if the planner selects actions in alphabetical order, it may continually choose A→B and then B→A, resulting in an endless A-B-A-B sequence. This pathological behavior illustrates how naïve DFS can fail in symbolic planners without mechanisms to detect or prevent cycles. On the other hand, using breadth-first search (BFS) can avoid such looping issues but may introduce inefficiencies due to exhaustive exploration.
This challenge raises a critical question: \textbf{How can the planner determine search strategies to accomplish tasks within constrained reasoning steps?}

To address this adaptation issue, \textbf{Dylan} employs a rule weight matrix \( \mathbf{W} = [ {\bf w}_1, \ldots, {\bf w}_M ] \) to dynamically select planning rules. By applying a \textit{softmax} function to each weight vector \( {\bf w}_j \in \mathbf{W} \), we choose \( M \) rules from a total of \( C \) rules. The weight matrix \( \mathbf{W} \) is initialized randomly and optimized via gradient descent, minimizing a Binary Cross-Entropy (BCE) loss between the target probability $p_{\text{target}}$ and the predicted probability $p_{\text{predicted}}$:
\begin{equation}
    \underset{\mathbf{W}}{\mathtt{minimize}} \quad  \mathtt{L_{loss}} = \mathtt{BCE}(p_{\text{target}},p_{\text{predicted}}(\mathbf{W})).
\end{equation}
Where the predicted probability $
    \mathbf{p}_{\text{predicted}} = \mathbf{v}^{(T)}\left[I_{\mathcal{G}}(\operatorname{target})\right]$, $
 I_{\mathcal{G}}(x)$ returns the index of the target atom within the set $ \mathcal{G} $.  
 \( \mathbf{v}^{(T)} \) denotes the valuation tensor obtained after \( T \)-step forward reasoning. \( \mathbf{v}[i] \) represents the \( i \)-th entry of the valuation tensor.  
Due to its differentiability, \textbf{Dylan} can dynamically adapt its search strategies based on the tasks, which we demonstrate in Section~\ref{experiments}.

\section{Experimental Evaluation}
\label{experiments}

In this section, we show how Dylan does 
reward shaping using human prior and its planning capability by composing different policy primitives. 
Overall, we aim to answer four research questions. \textbf{Q1:} Can Dylan enable reinforcement learning agents to learn more effectively with fewer interactions?
\textbf{Q2:} Can Dylan further improve RL agent learning performance by working as an adaptive reward model? 
\textbf{Q3:} Can Dylan compose policy primitives to generalize to a different task instead of retraining new policies? 
\textbf{Q4:} Can Dylan adapt itself to tasks that require different search strategies?

\textbf{Environment setup.} To evaluate and compare different reinforcement learning methods, we conduct a series of experiments within the MiniGrid environment suite \citep{chevalier2023minigrid}. MiniGrid offers a range of partially observable, grid-based tasks that are designed to test an agent’s generalization and exploration capabilities. For consistency and reproducibility, we focus on a subset of the MiniGrid-DoorKey environment (as shown in Fig.~\ref{fig:plannermodel}), in which the agent must first acquire a key, use it to open a door, and then navigate to the goal position. While the original MiniGrid-DoorKey environment offers only a single viable solution path, we have customized the  MiniGrid-DoorKey environments 
to support multiple solution paths. Each environment presents different exploration challenges and sparse reward settings, making them ideal benchmarks for evaluating learning performance.

\textbf{Baselines.} We compare the following reinforcement learning algorithms: \textbf{A2C~\citep{mnih2016asynchronous}:} Advantage Actor-Critic, a synchronous, on-policy policy gradient method known for its simplicity and stability. \textbf{PPO~\citep{schulman2017proximal}:} Proximal Policy Optimization , an on-policy actor-critic method with clipped objectives to ensure stable updates, which is the state of the art reinforcement learning algorithms. 

All methods are implemented using torch-ac~\citep{torch-ac}, unless otherwise specified, hyperparameters are selected from the literature without fine-tuning (training hyperparameters in Appendix~\ref{hyperparameter}).




\subsection{Dylan as reward model}

\begin{wrapfigure}{t}{0.5\textwidth}
    \centering
    \vspace{-0.3in}
    \includegraphics[trim=15 10 40 15, clip,width=0.40\textwidth]{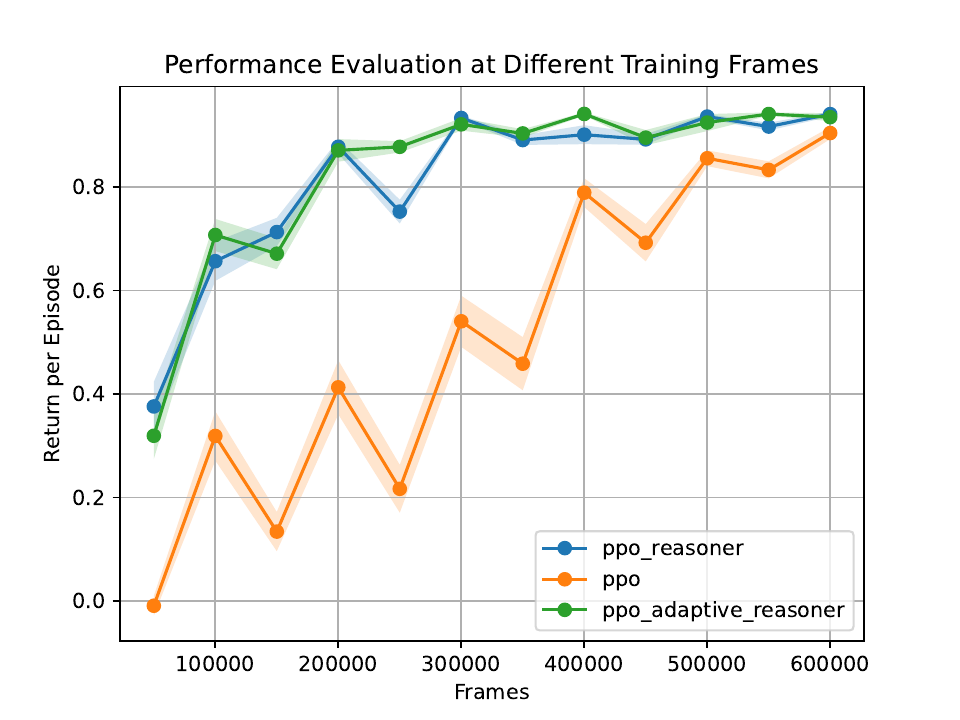}
    \caption{\textbf{Dylan further boost RL agents' performances as an adaptive reward model}, shown by the returns over training of the PPO baseline, and PPO with reasoner and adaptive reasoner, on the $\mathtt{8\times8}$ MiniGrid-DoorKey environment using static rewards, adaptive rewards, and pure PPO. Averages over $3$ runs; solid lines show means, shaded areas show standard error. (Best viewed in color)}
    \vspace{-0.2in}
    \label{ada-rewardmodel}
\end{wrapfigure}
In this experiment, we evaluate Dylan's effectiveness as a static reward model for guiding reinforcement learning agents during training. As shown in Fig.~\ref{fig:plannermodel}, the modified MiniGrid-DoorKey environment provides two solutions for the agent to reach the goal position (indicated by the green marker). The agent can either directly go through an already opened blue door or alternatively acquire a red key, use it to unlock a red door, and navigate to the goal.

The primary goal of our experiments is to quantify how Dylan influences the learning efficiency and convergence speed across different reinforcement learning algorithms. We conduct comparative evaluations using two benchmark MiniGrid-Doorkey environments: a moderately complex $\mathtt{12\times12}$ grid and a more challenging $\mathtt{16\times16}$ grid. Both environments emphasize the necessity for effective exploration due to their sparse reward setting. 

Fig.~\ref{fig:rewardmodel} presents a detailed comparison of training performance across several reinforcement learning algorithms, specifically PPO and A2C, evaluated both with and without Dylan’s auxiliary reward signals. Results are averaged over three independent runs; solid lines indicate mean performance, while shaded regions denote  the min and max values. To enhance readability, all curves are smoothed using a time-based exponential moving average (EMA).  Our empirical results clearly demonstrate Dylan's substantial positive impact on learning efficiency, particularly as the complexity of the environment increases. While all evaluated algorithms exhibit improved convergence rates when assisted by Dylan in the $\mathtt{12\times12}$ environment, the improvement becomes notably greater in the more demanding $\mathtt{16\times16}$ scenario. Remarkably, baseline agents such as PPO, which reliably converge in simpler settings, fail to converge in the challenging 
$\mathtt{16\times16}$ environment without Dylan's guidance. In contrast, incorporating Dylan effectively resolves these exploration bottlenecks, significantly accelerating convergence.
In summary, our findings strongly suggest that Dylan effectively mitigates common reinforcement learning challenges, such as sparse rewards and inefficient exploration strategies. Importantly, these improvements are achieved simply by adding Dylan as an auxiliary reward provider, without altering the original reinforcement learning algorithm.


\begin{wrapfigure}{t}{0.5\textwidth}
    \centering
   \vspace{-0.2in}
    \includegraphics[trim=15 10 10 10, clip, width=0.45\textwidth]{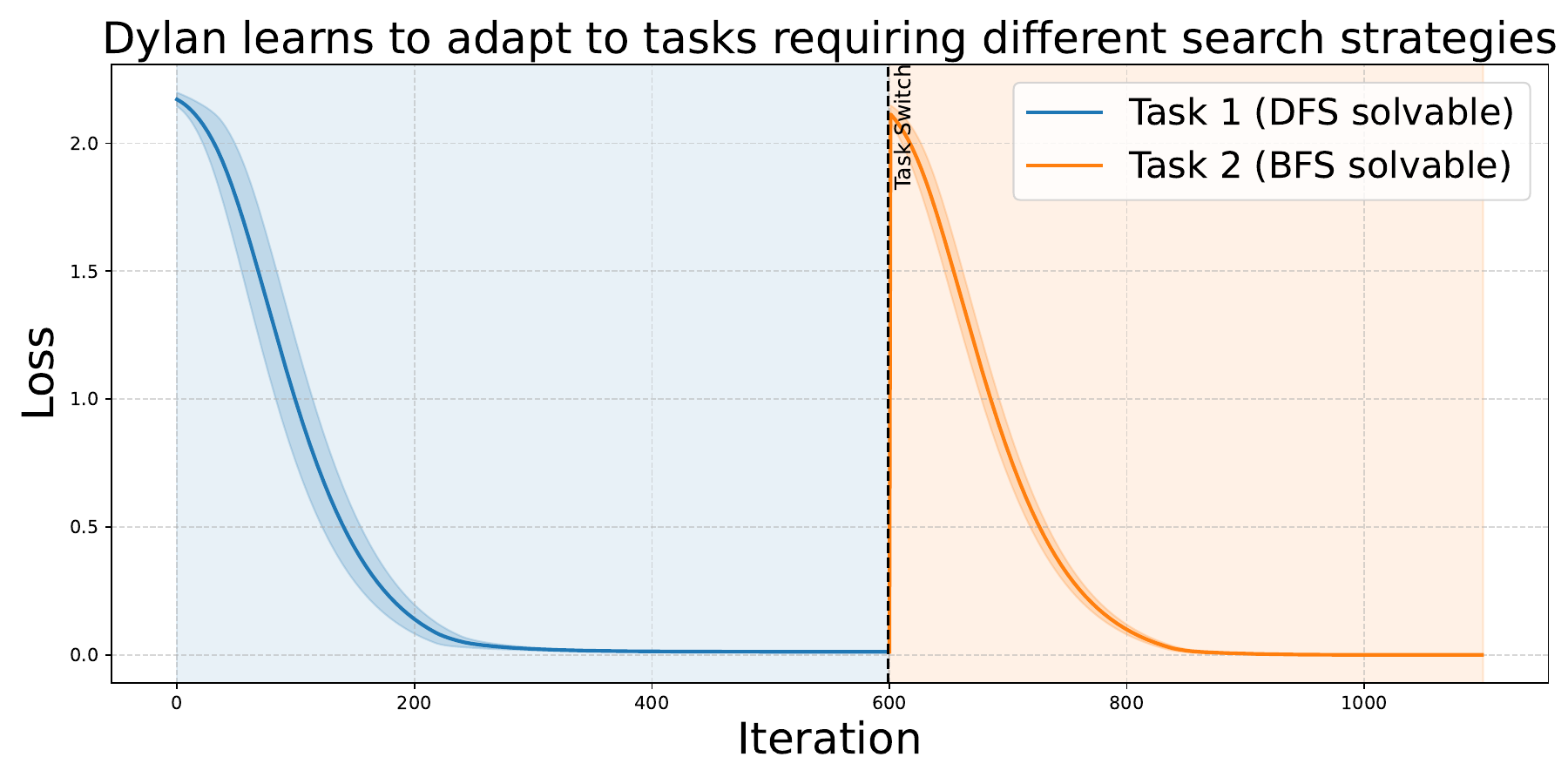}
    \caption{\textbf{Training loss curve as Dylan learns to adapt to multiple tasks requiring different search strategies.} Results are averaged over three runs, with solid lines indicating mean values and shaded areas representing standard deviation.  
    }
    \label{structurelearning}
    \vspace{-0.2in}
\end{wrapfigure}

\subsection{Dylan as adaptive reward model}
In this experiment, we further assess Dylan’s ability to guide the agent's learning by comparing two reward models (static and adaptive) alongside a baseline pure PPO approach. The static reward model delivers sparse rewards based on the agent’s immediate success, while the adaptive reward model provides denser and more informative rewards by dynamically evaluating the agent’s progress. 
Fig.~\ref{ada-rewardmodel} illustrates the learning performance of different methods in the MiniGrid-DoorKey $\mathtt{8 \times 8}$ environment, using both static and adaptive reward settings. Results are averaged over three independent runs, with solid lines indicating mean returns and shaded regions denoting standard error.

Our results show that incorporating Dylan’s auxiliary rewards consistently improves convergence speed compared to the pure PPO~\citep{schulman2017proximal} baseline. Notably, the adaptive reward model achieves slightly faster convergence than the static one, suggesting that accounting for multiple possible plans can further reduce training interactions.
It is important to note that we did not perform exhaustive hyperparameter tuning for the adaptive reward model. As such, the current configuration (hyperparameters in Appendix~\ref{hyperadaptive}) may not be optimal, and further tuning could potentially yield even greater improvements in learning efficiency.

\begin{table*}[t]
\small
\centering
\begin{tabular}{lcccc}
              & Key Retrieval  & Red Door Reaching & Goal Reaching & Safe Goal Reaching\\ 
A2C~\citep{mnih2016asynchronous}           &59.2 \textsubscript{$\pm$12.5}      &50.2\textsubscript{$ \pm$13.6}           & 98.6\textsubscript{$ \pm$ 1.6}    &41.2\textsubscript{$ \pm$ 17}            \\
PPO~\citep{schulman2017proximal}            &63.8\textsubscript{$ \pm$ 12.2}      &53.2\textsubscript{$ \pm$ 13.4}         & 100\textsubscript{$ \pm$ 0}  &42\textsubscript{$ \pm$ 17.2}          \\
Dylan~(ours)  &100\textsubscript{$ \pm$ 0}$\bullet$   &100\textsubscript{$ \pm$ 0} $\bullet$        &100\textsubscript{$ \pm$ 0} $\bullet$      &98.2\textsubscript{$ \pm$ 2.2} $\bullet$    \\ 
\end{tabular}
\caption{\textbf{Performance on multitasks setting (Successrate; the higher, the better).} We compare Dylan with the baseline method PPO~\citep{schulman2017proximal} and A2C~\citep{mnih2016asynchronous}. Success rates are averaged on fifty runs with its standard deviation. The best-performing models are denoted using $\bullet$.}
\label{tab:multi_tasks}
\vspace{-0.2in}
\end{table*}
\subsection{Dylan as differentiable planner}
In this experiment, we demonstrate Dylan’s ability to adapt its planning strategy to varying task structures and to compose policy primitives in order to generate novel behaviors. Specifically, we evaluate two core capabilities: adaptive search strategy selection and compositional generalization.
\paragraph{Dylan's adaptivity.} Certain scenarios may require different search strategies. For example, in the scenario (task in Appendix.~\ref{dfsfail}) Depth-First Search (DFS) may result in an infinite loop, where the agent aims to reach a state defined by $\mathtt{reach\_goal}$ by combining two policy primitives $\mathtt{go\_through\_red\_door}$ and $\mathtt{go\_through\_blue\_door}$. However, when employing DFS, the recursive definitions of $\mathtt{get\_through\_door}$ may result in repeated exploration of the same paths. For instance, the agent may continuously attempt to execute the $\mathtt{go\_through\_red\_door}$ or $\mathtt{go\_through\_blue\_door}$ actions without progressing towards the final goal. This is particularly problematic when the agent encounters cyclic rules or tasks with high branching factors, where DFS tends to prioritize depth exploration over breadth, thereby getting stuck in loops or excessively deep branches. Under such conditions, the agent should autonomously recognize the inefficiency of DFS and accordingly shift to a more suitable method, such as Breadth-First Search (BFS). In this task (task provided in Appendix~\ref{diff_plan}), the planner is asked to solve each task within three reasoning steps. The first task is best addressed using DFS, while the second is more efficiently solved with BFS. To perform well across both scenarios, the agent must dynamically adjust its planning strategy to suit the structure of each task. Fig.~\ref{structurelearning} shows the training loss curves for Dylan on these two representative tasks (planner rules provided in Appendix~\ref{diff_plan_rules}). All results are averaged over three runs, with the solid line representing the mean and the shaded areas indicating the standard deviation. 

\paragraph{Dylan's compositionality.} After demonstrating Dylan’s differentiable adaptivity, we further evaluate its ability to compose policy primitives to produce novel behaviors, enabling the agent to solve previously unseen tasks without retraining. In this experiment, the planner is provided with a set of reusable primitives, such as $\mathtt{get\_key}$ and $\mathtt{go\_through\_door}$, and is asked to solve a range of goal-directed tasks using these building blocks. We designed four tasks in this experiment. In these tasks, the agent must retrieve a key, navigate to the red door, and reach the goal position while avoiding the blue door, which leads to a trap. Note that we use the environment shown in Figure~\ref{fig:plannermodel}, where the agent has multiple possible paths to the goal. As a result, possessing a key is not necessarily a prerequisite for opening a door.
For comparison, we include PPO and A2C agents trained specifically to navigate to the goal. Table.~\ref{tab:multi_tasks} summarizes the various task settings within the MiniGrid-Doorkey environment. Dylan successfully composes low-level primitives to generalize across diverse tasks, whereas the PPO and A2C agents, which are trained solely for goal navigation, fail to generalize to tasks requiring more complex behavior.

Overall, the experimental results provide affirmative answers to all four questions \textbf{Q1}--\textbf{Q4}.
\section{Limitations}
\label{limitations}
Although Dylan has demonstrated impressive results in enabling agents to learn from less environment interactions and generalizing to new tasks, it also has certain limitations. One limitation lies in its reliance on symbolic states provided directly by the environment. In future work, we aim to explore the use of vision foundation models to extract symbolic representations directly from raw game images. Another limitation involves the generation of game rules by large language models (LLMs). Currently, we prompt the GPT-4o to produce game rules and rely on human supervision to verify and correct them. A promising direction for future research is to incorporate an automated error-correction mechanism—such as leveraging multiple LLMs in a multi-round discussion framework—to improve the accuracy and reliability of the generated rules.

\section{Conclusions}
\label{conclusions}

In this paper, we introduced Dylan, a novel reward-shaping framework that leverages human prior knowledge to help reinforcement learning agents learn with less training interactions. Beyond its role as a reward model, Dylan is, to the best of the authors' knowledge, the first differentiable symbolic planner that alleviates traditional symbolic planner's non-adaptable limitation. Dylan is capable of dynamically combining primitive policies to synthesize novel, complex behaviors. Our empirical evaluations demonstrate Dylan’s effectiveness in enabling agents to learn from fewer interactions, accelerating convergence, and overcoming exploration bottlenecks—especially in environments with increasing complexity. These results illustrate Dylan’s potential as a robust framework bridging symbolic reasoning and reinforcement learning.

Promising avenues for future research include automating the acquisition of symbolic abstractions, for example, through predicate invention. Additionally, investigating Dylan’s scalability and broader applicability across both symbolic and subsymbolic domains remains an interesting research direction. We further want to apply Dylan to multimodal scenarios, where it could leverage multimodal inputs to more effectively guide agent learning across diverse and complex environments.
\section{Acknowledgements}
\label{Acknowledgements}
This work is supported by the Hessian Ministry of Higher Education, Research, Science and the Arts
(HMWK; projects ``The Third Wave of AI'' and ``The Adaptive Mind''). It further benefited from the
Hessian research priority programme LOEWE within the project ``WhiteBox'' and the EU-funded
``TANGO'' project (EU Horizon 2023, GA No 57100431) and  Federal Ministry of Education and Research under the funding code 01|S25002B. Furthermore, the authors would like to thank Quentin Delfosse, Cedric Derstroff and Jannis Brugger for proofreading our paper and thus improving the clarity of the final manuscript.

\bibliography{neurips_2025}
\bibliographystyle{plain}

\newpage
\section*{NeurIPS Paper Checklist}
\begin{enumerate}

\item {\bf Claims}
    \item[] Question: Do the main claims made in the abstract and introduction accurately reflect the paper's contributions and scope?
    \item[] Answer: \answerYes{} 
    \item[] Justification: We claim in the abstract and introduction that we integrate a differentiable symbolic planner with reinforcement learning for computing a shaped reward function, or as high-level policy in a hierarchical setting, and thereby increase efficiency during training. We discuss this integration in Section~\ref{dylan} and demonstrate the improved efficiency in Section~\ref{experiments}.

\item {\bf Limitations}
    \item[] Question: Does the paper discuss the limitations of the work performed by the authors?
    \item[] Answer: \answerYes{} 
    \item[] Justification: in Sec.~\ref{limitations} we discuss the limitations of our work

\item {\bf Theory assumptions and proofs}
    \item[] Question: For each theoretical result, does the paper provide the full set of assumptions and a complete (and correct) proof?
    \item[] Answer: \answerNA{} 
    \item[] Justification: We do not include theoretical results.

    \item {\bf Experimental result reproducibility}
    \item[] Question: Does the paper fully disclose all the information needed to reproduce the main experimental results of the paper to the extent that it affects the main claims and/or conclusions of the paper (regardless of whether the code and data are provided or not)?
    \item[] Answer: \answerYes{} 
    \item[] Justification: The paper fully discloses all necessary details to reproduce the main experimental results. This includes descriptions of the experimental setup~\ref{experiments}, methods~\ref{experiments} and hyperparameters~\ref{hyperparameter}. All information critical to supporting the main claims and conclusions is explicitly stated, ensuring reproducibility even without direct access to the code or data.

\item {\bf Open access to data and code}
    \item[] Question: Does the paper provide open access to the data and code, with sufficient instructions to faithfully reproduce the main experimental results, as described in supplemental material?
    \item[] Answer: \answerYes{} 
    \item[] Justification: We include our source code in to supplemental material.

\item {\bf Experimental setting/details}
    \item[] Question: Does the paper specify all the training and test details (e.g., data splits, hyperparameters, how they were chosen, type of optimizer, etc.) necessary to understand the results?
    \item[] Answer: \answerYes{} 
    \item[] Justification: We discuss the experimental setting in Section~\ref{experiments} and present the hyperparameters in Appendix~\ref{hyperparameter}.

\item {\bf Experiment statistical significance}
    \item[] Question: Does the paper report error bars suitably and correctly defined or other appropriate information about the statistical significance of the experiments?
    \item[] Answer: \answerYes{} 
    \item[] Justification: The paper reports error bars appropriately. For each experiment, we indicate the number of runs and specify whether the shaded regions represent standard deviation or standard error. This ensures clarity and transparency in the statistical significance of the reported results.

\item {\bf Experiments compute resources}
    \item[] Question: For each experiment, does the paper provide sufficient information on the computer resources (type of compute workers, memory, time of execution) needed to reproduce the experiments?
    \item[] Answer: \answerYes{} 
    \item[] Justification: Our computational requirements are modest and can be met by any standard desktop. All experiments were run on an M1 Pro CPU with 16GB of memory, without using a GPU. 
    
\item {\bf Code of ethics}
    \item[] Question: Does the research conducted in the paper conform, in every respect, with the NeurIPS Code of Ethics \url{https://neurips.cc/public/EthicsGuidelines}?
    \item[] Answer: \answerYes{} 
    \item[] Justification: All authors read and adhere to the NeurIPS Code of Ethics.

\item {\bf Broader impacts}
    \item[] Question: Does the paper discuss both potential positive societal impacts and negative societal impacts of the work performed?
    \item[] Answer: \answerNA{} 
    \item[] Justification: While there is always potential for positive or negative societal impact, we could not identify a potential consequence that is worth highlighting.
    
\item {\bf Safeguards}
    \item[] Question: Does the paper describe safeguards that have been put in place for responsible release of data or models that have a high risk for misuse (e.g., pretrained language models, image generators, or scraped datasets)?
    \item[] Answer: \answerNA{} 
    \item[] Justification: We do not release a pre-trained model or data set.

\item {\bf Licenses for existing assets}
    \item[] Question: Are the creators or original owners of assets (e.g., code, data, models), used in the paper, properly credited and are the license and terms of use explicitly mentioned and properly respected?
    \item[] Answer: \answerYes{} 
    \item[] Justification: All external assets used in this paper—including code, datasets, and models—are properly credited and are licensed under the CC-BY 4.0 license

\item {\bf New assets}
    \item[] Question: Are new assets introduced in the paper well documented and is the documentation provided alongside the assets?
    \item[] Answer: \answerYes{} 
    \item[] Justification: Our code is provided with a readme file.

\item {\bf Crowdsourcing and research with human subjects}
    \item[] Question: For crowdsourcing experiments and research with human subjects, does the paper include the full text of instructions given to participants and screenshots, if applicable, as well as details about compensation (if any)? 
    \item[] Answer: \answerNA{} 
    \item[] Justification: We did not perform crowdsourcing or human experiments for this project.

\item {\bf Institutional review board (IRB) approvals or equivalent for research with human subjects}
    \item[] Question: Does the paper describe potential risks incurred by study participants, whether such risks were disclosed to the subjects, and whether Institutional Review Board (IRB) approvals (or an equivalent approval/review based on the requirements of your country or institution) were obtained?
    \item[] Answer: \answerNA{} 
    \item[] Justification: We did not perform human experiments for this project.

\item {\bf Declaration of LLM usage}
    \item[] Question: Does the paper describe the usage of LLMs if it is an important, original, or non-standard component of the core methods in this research? Note that if the LLM is used only for writing, editing, or formatting purposes and does not impact the core methodology, scientific rigorousness, or originality of the research, declaration is not required.
    \item[] Answer: \answerYes{} 
    \item[] Justification: We used an LLM to automatically extract the game rules for the symbolic planner. Details are given in the appendix.

\end{enumerate}
\newpage
\appendix
\section{Symbolic transformation for Dylan}
\label{prompt}
Prompt: read the game manual and return the game rules in first-oder logic format such as:
\begin{align*}
    &\mathtt{get\_through\_door} \textbf{:- } \mathtt{initial, go\_through\_red\_door.} \\
    &\mathtt{get\_through\_door} \textbf{:- } \mathtt{get\_through\_door,} \mathtt{go\_through\_blue\_door.}\\
    &\mathtt{get\_through\_door} \textbf{:- } \mathtt{get\_through\_door,} \mathtt{go\_through\_red\_door.}\\
    &\mathtt{reach\_goal}\textbf{ :- } \mathtt{get\_through\_door,}\mathtt{go\_to\_goal.}
\end{align*}

Dylan leverages human prior knowledge to guide reinforcement learning. As the first step, this prior knowledge is converted into a structured, logic-based format. For example, consider the agent navigation task illustrated in Fig.~\ref{fig:plannermodel}, drawn from MiniGrid Doorkey environment \citep{chevalier2023minigrid}, After reviewing the environment’s manual, we extract a rule stating that a locked door can only be opened with a key of the corresponding color, and that the agent can reach the goal by passing through an opened door. This rule can be expressed as the logic program: 
\begin{align*}
&\mathtt{get\_red\_key}\textbf{:-}
\mathtt{init,go\_red\_key.}\qquad
\mathtt{go\_through\_door}\textbf{:-}\mathtt{init,}\mathtt{go\_blue\_door.}\\
&\mathtt{go\_through\_door}\textbf{:-}\mathtt{get\_red\_key,}\mathtt{go\_open\_red\_door.}\\
&\mathtt{reach\_goal}\textbf{:-}
\mathtt{go\_through\_door,}\mathtt{go\_to\_goal.}
\end{align*}
We categorize the atoms in these clauses into two types: \textbf{state atoms} and \textbf{policy atoms}. Within each clause, the \textbf{head} and the \textbf{first body atom} are considered state atoms, while the \textbf{second body atom} corresponds to a policy atom. Following the STRIPS-style representation~\citep{fikes1971strips,fikes1993strips}, we transform each clause into a single atom using the predicate: $$
\mathtt{move/3/[action, pre\_condition, post\_condition]}. $$
For example, the clause: $
\mathtt{get\_blue\_key} \textbf{ :- } \mathtt{initial, go\_blue\_key}.
$
is rewritten as one atom:
$
\mathtt{move(go\_blue\_key, initial, get\_blue\_key)}.
$
In this transformation, the \textbf{head} (
$\mathtt{get\_blue\_key}$) becomes the \textbf{postcondition}, the \textbf{first body atom} (
$\mathtt{initial}$) serves as the \textbf{precondition}, and the \textbf{second body atom} (
$\mathtt{go\_blue\_key}$) represents the \textbf{action}.

With this transformation at hand, we now a symbolic planner:
\begin{align*}
    &\mathtt{plan(Start, New, Goal, [Act, Old\_stack])} \textbf{ :- } \\
    &\hspace{10ex} \mathtt{move(Act, Old, New),}\mathtt{condition\_met(Old, Current),}\\
    &\hspace{10ex} \mathtt{change\_state(Current, New),}\mathtt{plan(Start, Current, Goal, Old\_stack).} \\
    &\mathtt{plan\_final(Start, Goal, Move\_stack)} \textbf{ :- } \\
    &\hspace{10ex} \mathtt{plan(Start, Current, Goal, Move\_stack),}\mathtt{equal(Current, Goal).}
\end{align*}
The first rule establishes the recursive process for generating a plan. It selects an appropriate policy that transitions the agent from an old state to a new state, verifies the necessary conditions, updates the current state accordingly, and recursively continues the planning process until the goal is reached. The second rule defines the termination condition, ensuring that the planning process concludes once the agent's current state matches the desired goal state. 
\section{Differentialize Symbolic planner}
\label{diff}
We now describe each step in detail on how to differentialize the symbolic planner.

\paragraph{(Step 1) Encoding Logic Programs as Tensors.}
To enable differentiable forward reasoning, Each meta-rule is transformed into a tensor representation for differentiable forward reasoning. Each meta-rule $C_i \in \mathcal{C}$ is encoded as a tensor $\mathbf{I}_i \in \mathbb{N}^{G \times S \times L}$, where $S$ denotes the maximum number of substitutions for existentially quantified variables in the rule set , and $L$ is the maximum number of atoms in the body of any rule. 
For instance, $\mathbf{I}_i[j, k, l]$ stores the index of the $l$-th subgoal in the body of rule $C_i$ used to derive the $j$-th fact under the $k$-th substitution.



\paragraph{(Step 2) Weighting and Selecting Meta-Rules.}
We construct the reasoning function by assigning weights that determine how multiple meta-rules are combined.
(i) We fix the size of the target meta-program to be $M$, meaning the final program will consist of $M$ meta-rules selected from a total of $C$ candidates in $\mathcal{C}$.
(ii) To enable soft selection, we define a weight matrix $\mathbf{W} = [\mathbf{w}_1, \ldots, \mathbf{w}_M]$, where each $\mathbf{w}_i \in \mathbb{R}^C$ assigns a real-valued weight.
(iii) We then apply a \emph{softmax} to each $\mathbf{w}_i$ to obtain a probability distribution over the $C$ candidates, allowing the model to softly combine multiple meta-rules. 
\paragraph{(Step 3) Perform Differentiable Inference.}
Starting from a single apply of the weighted meta-rules, we iteratively propagate inferred facts across $T$ reasoning steps. 


We compute the valuation of body atoms for every grounded instance of a meta-rule $C_i \in \mathcal{C}$. This is achieved by first gathering the current truth values from the valuation vector $\mathbf{v}^{(t)}$ using an index tensor $\mathbf{I}i$, and then applying a multiplicative aggregation across subgoals:
\begin{align}
b_{i,j,k}^{(t)} = \prod_{l=1}^{L} \mathbf{gather}(\mathbf{v}^{(t)}, \mathbf{I}i)[j,k,l],
\label{eq:gather_prod_body}
\end{align}
where the $\mathbf{gather}$ operator maps valuation scores to indexed body atoms:
\begin{align}
\mathbf{gather}(\mathbf{x}, \mathbf{Y})[j,k,l] = \mathbf{x}[\mathbf{Y}[j,k,l]].
\end{align}
The resulting value $b_{i,j,k}^{(t)} \in [0,1]$ reflects the conjunction of subgoal valuations under the $k$-th substitution of existential variables, used to derive the $j$-th candidate fact from the $i$-th meta-rule. Logical conjunction is implemented via element-wise product, modeling the "and" over the rule body.


To integrate the effects of multiple groundings of a meta-rule $C_i$, we apply a smooth approximation of logical \emph{or} across all possible substitutions. Specifically, we compute the aggregated valuation $c^{(t)}_{i,j} \in [0,1]$ as:
\begin{align}
c^{(t)}_{i,j} = \mathit{softor}^\gamma(b_{i,j,1}^{(t)}, \ldots, b_{i,j,S}^{(t)}),
\end{align}
where $\mathit{softor}^\gamma$ denotes a differentiable relaxation of disjunction. This operator is defined as:
\begin{align}
\mathit{softor}^\gamma(x_1, \ldots, x_n) = \gamma \log \sum_{i=1}^{n} \exp(x_i / \gamma),
\label{eq:softor}
\end{align}
with temperature parameter $\gamma > 0$ controlling the smoothness of the approximation. This formulation closely resembles a softmax over valuations and serves as a continuous surrogate for the logical \emph{max}, following the log-sum-exp technique commonly used in differentiable reasoning \citep{cuturi2017soft}.


\textbf{(ii) Weighted Aggregation Across Meta-Rules.}
We compute a weighted combination of meta rules using the learned soft selections:
\begin{align}
h_{j,m}^{(t)} = \sum_{i=1}^{C} w^{m,i} \cdot c^{(t)}{i,j},
\end{align}
where $h^{(t)}_{j,m} \in [0,1]$ represents the intermediate result for the $j$-th fact contributed by the $m$-th slot. Here, $w^{m,i}$ is the softmax-normalized score over the $i$-th meta-rule:
\begin{align*}
w^*_{m,i} = \frac{\exp(w_{m,i})}{\sum_{i'} \exp(w_{m,i'})}, \quad w_{m,i} = \mathbf{w}_m[i].
\end{align*}

Finally, we consolidate the outputs of the $M$ softly selected rule components using a smooth disjunction:
\begin{align}
r_j^{(t)} = \mathit{softor}^\gamma(h^{(t)}_{j,1}, \ldots, h^{(t)}_{j,M}),
\label{eq:softor_weighted_rules}
\end{align}
which yields the $t$-step valuation for fact $j$. This mechanism allows the model to integrate $M$ soft rule compositions from a larger pool of $C$ candidates in a fully differentiable way.


\textbf{(iii) Iterative Forward Reasoning.}
We iteratively apply the forward reasoning procedure for $T$ steps. At each step $t$, we update the valuation of each fact $j$ by softly merging its newly inferred value $r_j^{(t)}$ with its previous valuation:
\begin{align}
v^{(t+1)}_j = \mathit{softor}^\gamma(r_j^{(t)}, v_j^{(t)}).
\label{eq:softor_weighted_rules}
\end{align}
This recursive update mechanism approximates logical entailment in a differentiable form, enabling the model to perform $T$-step reasoning over the evolving fact valuations.
The whole reasoning computation Eq.~\ref{eq:gather_prod_body}-\ref{eq:softor_weighted_rules} can be implemented using efficient tensor operations.
\section{Hyperparameter}
\label{hyperparameter}
\begin{table}[ht]
\caption{Summary of Training Hyperparameters}
\centering
\begin{tabular}{ll}
\toprule
\textbf{Parameter} & {\textbf{Training Parameters}} \\
\midrule
\texttt{--epochs} & 4 (PPO optimization epochs per update) \\
\texttt{--batch-size} & 256 (Batch size for PPO updates) \\
\texttt{--frames-per-proc} & 128 (Frames per process before update) \\
\texttt{--discount} & 0.99 (Discount factor $\gamma$) \\
\texttt{--lr} & 0.0001 (Learning rate) \\
\texttt{--gae-lambda} & 0.95 ($\lambda$ for GAE) \\
\texttt{--entropy-coef} & 0.01 (Entropy regularization coefficient) \\
\texttt{--value-loss-coef} & 0.5 (Value loss coefficient) \\
\texttt{--max-grad-norm} & 0.5 (Gradient clipping norm) \\
\texttt{--optim-eps} & $1\times10^{-8}$ (Optimizer epsilon) \\
\texttt{--optim-alpha} & 0.99 (RMSprop alpha) \\
\texttt{--clip-eps} & 0.2 (PPO clipping parameter $\epsilon$) \\
\texttt{--recurrence} & 1 (Recurrent steps, LSTM if $>1$) \\
\texttt{--text} & \texttt{False} (Enable GRU for text input) \\
\bottomrule
\end{tabular}
\label{tab:hyperparams}
\end{table}

\section{Differentiable Planning Task}
\label{diff_plan}
In this appendix, we present two pathological tasks that require a planner to adapt its search strategy—specifically between Depth-First Search (DFS) and Breadth-First Search (BFS)—for efficient goal finding within three steps. These examples demonstrate how a fixed strategy may underperform depending on the search structure and goal location.

\textbf{Task 1} favors DFS, as the goal \texttt{plan(a, h)} lies deep along a single branch:
\begin{verbatim}
edge(a, c). edge(a, b). edge(a, d).
edge(b, e). edge(c, f). edge(d, g).
edge(e, h).
\end{verbatim}

\textbf{Task 2} favors BFS, as the goal \texttt{plan(a, e)} is close to the root but may be delayed by DFS exploring deeper branches first:
\begin{verbatim}
edge(a, c). edge(a, b).edge(b, d).
edge(c, e). edge(d, f).
\end{verbatim}

\section{Differentiable Planning rules}
\label{diff_plan_rules}
We define the planning behavior of two search algorithms—\textbf{DFS} and \textbf{BFS}—using logical rules. These rules describe how each algorithm explores the search space and identifies successful plans.
\begin{align*}
&\texttt{dfs(B, F, G, r(F, D))} :- \texttt{edge(E, F), dfs(B, E, G, D)}. \\
&\texttt{plan(B, G)} :- \texttt{dfs(B, F, G, H), equal(F, G)}. \\
&\texttt{bfs(k(B, D), S, E)} :- \texttt{findall(A, F), append(C, F, k(B, D)), bfs(k(A, C), S, E)}. \\
&\texttt{plan(S, E)} :-\texttt{bfs(k(A, C), S, E), equalbfs(A, C, E)}.
\end{align*}

\section{Task that DFS can fail}
\label{dfsfail}
\begin{align*}
    &\mathtt{get\_through\_door} \textbf{:- } \mathtt{initial, go\_through\_red\_door.} \\
    &\mathtt{get\_through\_door} \textbf{:- } \mathtt{get\_through\_door,} \mathtt{go\_through\_blue\_door.}\\
    &\mathtt{get\_through\_door} \textbf{:- } \mathtt{get\_through\_door,} \mathtt{go\_through\_red\_door.}\\
    &\mathtt{reach\_goal}\textbf{:- } \mathtt{get\_through\_door,}\mathtt{go\_to\_goal.}
\end{align*}
\section{First-Oder Logic}
\label{FOL}

In first-order logic, a term can be a constant, a variable, or a function term constructed using a function symbol. We denote an $n$-ary predicate ${\tt p}$ as ${\tt p}/(n, [{\tt dt_1}, \ldots, {\tt dt_n}])$, where ${\tt dt_i}$ represents the data type of the $i$-th argument.
An atom is an expression of the form ${\tt p(t_1, \ldots, t_n)}$, where ${\tt p}$ is an $n$-ary predicate symbol and ${\tt t_1}, \ldots, {\tt t_n}$ are terms. If the atom contains no variables, it is referred to as a ground atom, or simply a fact.

A literal is either an atom or the negation of an atom. We refer to an atom as a positive literal, and its negation as a negative literal.
A clause is defined as a finite disjunction ($\lor$) of literals. When a clause contains no variables, it is called a ground clause.
A definite clause is a special case: a clause that contains exactly one positive literal. Formally, if $A, B_1, \ldots, B_n$ are atoms, then the expression $A \lor \lnot B_1 \lor \ldots \lor \lnot B_n$ constitutes a definite clause. We write definite clauses in the form of $A~\mbox{:-}~B_1,\ldots,B_n$. where $A$ is the {\it head} of the clause, and the set $\{B_1, \ldots, B_n\}$ is referred to as the body. For simplicity, we refer to definite clauses as clauses throughout this paper. The forward-chaining inference is a type of inference in first-order logic to compute logical entailment~\citep{Russel09}.
\section{initial valuation obtained for adaptive reward model}
\label{v0}
To obtain the initial valuation \( v_0 \) for the adaptive reward model, we initialize both the environment's logic state and the high-level action atoms. Since the environment logic states are externally provided, the only variable component is the valuation of the high-level action atoms. We assign these valuations based on the agent’s distance to each subtask, using the inverse of the distance (plus a small positive constant) to produce a meaningful and smoothly varying probability. This design ensures that as the distance changes, the corresponding probability in the plan is updated accordingly.

In our adaptive reward model experiments, we define the valuation of each high-level action atom as \( 0.5 + \frac{1}{\text{distance} + 2} \). The plan probability is then computed based on the initial valuation of both the action atoms and the environment atoms.

\section{hyperparameters for adaptive reward model}
\label{hyperadaptive}
$\lambda = 0.01 \qquad \omega = 1/20$

\end{document}